\documentclass[10pt,twocolumn,letterpaper]{article}

\usepackage[pagenumbers]{cvpr} 
\usepackage{graphicx}
\usepackage{amsmath}
\usepackage{amssymb}
\usepackage{booktabs}
\usepackage{caption}
\usepackage{subcaption}
\usepackage[table,xcdraw]{xcolor}
\usepackage[pagebackref,breaklinks,colorlinks]{hyperref}

\usepackage[capitalize]{cleveref}
\crefname{section}{Sec.}{Secs.}
\Crefname{section}{Section}{Sections}
\Crefname{table}{Table}{Tables}
\crefname{table}{Tab.}{Tabs.}

\begin{document}

\title{Track Anything Rapter(TAR)}

\author{Fnu Obaid ur Rahman\\
Maryland Applied Graduate Engineering\\
University of Marlyand\\
{\tt\small obdurhmn@umd.edu}
\and
Tharun Puthanveettil\\
Maryland Applied Graduate Engineering\\
University of Marlyand\\
{\tt\small tvpian@umd.edu}
\thanks{\href{https://github.com/tvpian/Project-TAR}{https://github.com/tvpian/Project-TAR}}
}

\maketitle

\begin{abstract}
Object tracking is a fundamental task in computer vision with broad practical applications across various domains, including traffic monitoring, robotics, and autonomous vehicle tracking. In this project, we aim to develop a sophisticated aerial vehicle system known as Track Anything Rapter (TAR), designed to detect, segment, and track objects of interest based on user-provided multimodal queries, such as text, images, and clicks. TAR utilizes cutting-edge pre-trained models like DINO, CLIP, and SAM to estimate the relative pose of the queried object. The tracking problem is approached as a Visual Servoing task, enabling the UAV to consistently focus on the object through advanced motion planning and control algorithms. We showcase how the integration of these foundational models with a custom high-level control algorithm results in a highly stable and precise tracking system deployed on a custom-built PX4 Autopilot-enabled Voxl2 M500 drone. To validate the tracking algorithm's performance, we compare it against Vicon-based ground truth. Additionally, we evaluate the reliability of the foundational models in aiding tracking in scenarios involving occlusions. Finally, we test and validate the model's ability to work seamlessly with multiple modalities, such as click, bounding box, and image templates.
\end{abstract}

\section{Introduction}
\label{sec:intro}
This project aims to design and implement a system that utilizes state-of-the-art (SOTA) pre-trained models to accurately detect and track the pose of any target object through multimodal queries, such as clicks, bounding boxes, and images, using a custom-built UAV. 

Existing robotic systems for object detection and tracking exhibit two prominent limitations. Firstly, they are closed-set systems, meaning that they operate under the assumption that the set of objects to be detected and tracked is predetermined during the training phase \cite{nousi2019embedded}\cite{zhao2022vision}\cite{ganti2016implementation}\cite{unlu2019deep}. Consequently, these systems can only handle a fixed set of object categories. This characteristic restricts their adaptability since accommodating new object categories requires fine-tuning the model. Secondly, the objects of interest are typically identified solely by a class label, which may not be intuitive for end-users to specify. This constraint imposes limitations on how users can interact with the system, potentially hindering its usability and flexibility.\cite{maalouf2023deep}\cite{barisic2019vision}\cite{bartak2015any}

\begin{figure}[!htbp]
     \centering
    \includegraphics[width=\linewidth]{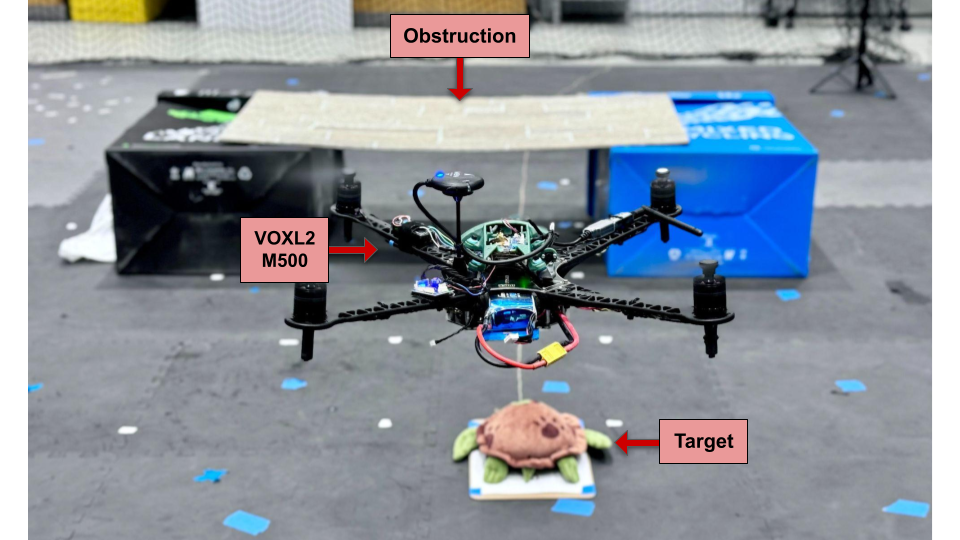}
    \caption{Experimental setup used for testing the performance of the one-shot model-DINO in aiding in redetection of targets in case of obstructions.}
    \label{fig:experimental_setup}

    \includegraphics[width=\linewidth]{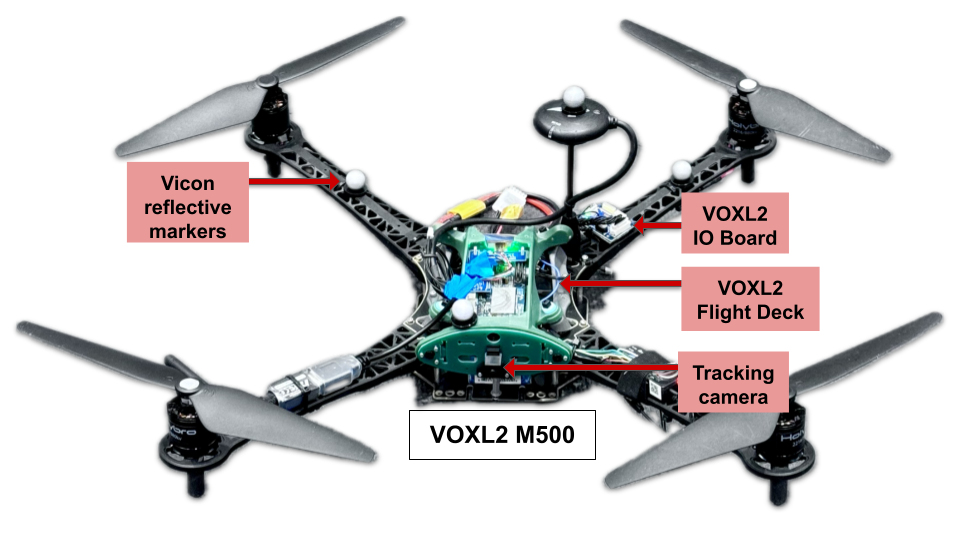}
    \caption{Rapter: The custom-built drone used for the experiments}
    \label{fig:rapter}
    
    \vspace{0.5cm} 
    
\end{figure}

Our research draws inspiration from the work presented in \cite{maalouf2024follow}, wherein utility functions and multi-modal model integration techniques were extensively employed by the authors. We have leveraged and adapted these methodologies to suit our objectives, particularly focusing on compatibility with our PX4-enabled custom-built drone. Additionally, we have developed a custom high-level algorithm tailored to effectively utilize the outputs of the multi-modal models for precise object tracking.

To evaluate the robustness and efficacy of the stated detection and tracking models, we introduced obstructions to simulate real-world scenarios. This allowed us to assess the models' performance in feature extraction and re-identification of the target object amidst potential distractors within the frame. Furthermore, we conducted experiments to assess the models' capability to handle multi-modal input queries and facilitate efficient tracking of the target object as stated by the authors using our custom-built drone.

\subsection{Contribution}
In this work, we make four significant contributions that extend the efforts of the authors of the base paper:

\begin{enumerate}
    \item \textbf{Custom Evaluation Pipeline:} We designed and implemented a bespoke pipeline to rigorously evaluate the performance of the proposed algorithms in the base paper in ROS2 Gazebo simulation, enabling a thorough assessment and validation of their efficacy.
    \item \textbf{Proportional-Based High-Level Controller:} We developed a custom proportional-based high-level controller to address the problem as a visual servoing task, effectively leveraging the low-level control capabilities provided by the PX4 Autopilot.
    \item \textbf{ROS2-Based Implementation:} We implemented an equivalent ROS2-based system utilizing MAVROS, thereby bypassing the need for MAVSDK proposed in the base paper for drone control, facilitating enhanced integration and operational flexibility.
    \item \textbf{DTW-based Trajectory evaluation:}Finally, we discussed the utilization of standard metrics such as Discrete Time Warping (DTW) to evaluate the trajectory traced by the drone during object tracking. This approach provides a systematic method for assessing the quality and effectiveness of the proposed tracking algorithm. By employing DTW, we can quantitatively measure the similarity between the drone's trajectory and the expected path, providing valuable insights into the performance and accuracy of the tracking system.
    
\end{enumerate}

\section{Methodology}
The implementation of the one-shot multi-modal tracking system involves a series of structured steps. This system incorporates state-of-the-art Vision Transformer (ViT) models, optimizing them for real-time performance, and integrates them into a cohesive framework. Specifically, we utilize the Segment Anything Model (SAM)\cite{Kirillov_2023_ICCV} for segmentation and DINO\cite{zhang2022dino} for general-purpose visual feature extraction. An efficient detection and semantic segmentation scheme is achieved by combining the features from DINO with the class-agnostic instance segmentation provided by SAM, as proposed in the base paper. For real-time tracking, we employ the (Seg)AOT\cite{bartak2015any} model and have designed a lightweight ROS-based visual servoing controller to ensure precise object tracking and following.

\subsection{Detection}
In our pipeline, we utilize SAM and DINO models for object detection and segmentation. The input comprises an RGB frame \( F \) from a video stream and a query \( q \), which can be in the form of an image template, click, or a bounding box on the input image.

Firstly, the segmentation operator \( Seg(F) \) is applied to \( F \), producing a set of masks \( \{M_1, \dots, M_n\} \). These masks partition the frame into distinct objects or regions, with each mask \( M_i \) representing a binary matrix denoting the segmented object's pixels.

Next, the feature extractor model \( Desc \) is employed to derive descriptors from \( F \). These descriptors capture semantic information about each pixel in \( F \), resulting in a descriptor tensor \( D \) with dimensions \( h \times w \times d \), where \( h \) and \( w \) represent the height and width of the frame respectively, and \( d \) is the dimensionality of the descriptor vectors.

To detect the desired object described by the query \( q \), we compute the feature descriptor \( v \) for \( q \) using \( Desc(q) \). This feature descriptor encapsulates information about the object described by \( q \) in the feature space.

Now, for each frame \( F_i \) received from the video stream, the following steps are executed:

\begin{enumerate}
    \item Instance Segmentation: \( Seg \) is applied to \( F_i \) to compute the segmentation masks \( \{M_1, \dots, M_n\}_i \), effectively partitioning the frame into regions. However, these regions are not yet classified as identified objects, and some regions may intersect.
    \item Descriptor Extraction: \( Desc \) is then applied to \( F_i \) to extract pixel-wise descriptors \( D_i \).
    \item Aggregation: Per-pixel descriptors are aggregated to form region-level descriptors. This aggregation is performed using average pooling, resulting in mean feature descriptors \( v_j \) for each segmentation region \( j \).
    \item Similarity Computation: The cosine similarity between each region descriptor \( v_j \) and the query feature descriptor \( v \) is computed. If the similarity exceeds a threshold, the region is labeled with the query, indicating that it corresponds to the desired object.
\end{enumerate}

In cases where multiple queries are provided, the most similar query is assigned to each region based on cosine similarity scores.


\subsection{Re-detecting a lost object}
Three re-detection methods are offered for temporary object loss during tracking, catering to different needs. The system automatically initiates re-detection when needed, and users can choose the level of support before starting the FAn(base paper pipeline): The first level relies on the tracker to re-detect the object, it’s the fastest and less robust, occasionally leading to false detections of similar objects. The second approach involves human-in-the-loop re-detection, requiring a user to click/draw a bounding box when tracking is lost, assuming human availability, which isn’t always possible. To mitigate this, we also propose an automatic re-detection technique.

Automatic re-detection via cross-trajectory stored ViT features:
To enable a robust and accurate autonomous re-detection of the tracked (lost) object, we provide a feature-descriptor storing mechanism for the tracked object in different stages of the tracking process, these stored features will be used to find the object once lost. Specifically, we suggest the following. Let \( \tau > 0 \) be an integer. During the tracking, at each iteration \( i \) such that \( i \mod \tau = 0 \), define \( M_{obj}^i \) to be the mask denoting the current tracked object in the frame, we first apply \( Desc \) on the current frame \( F_i \) to obtain \( D_i := Desc(F_i) \in \mathbb{R}^{h \times w \times d} \), then we compute the mean descriptor of the current tracked object as:
\[ v_{obj}^i := \frac{1}{\text{non-zero}(M_{obj}^i)} \sum_{p \in D_i[M_{obj}^i]} p \]
and add it to the set of previously computed descriptors to obtain the set \( V_{obj} := \{ v_{obj}^0 , v_{obj}^{\tau} , v_{obj}^{2\tau} , \dots , v_{obj}^{i} \} \). Whenever the system loses the tracked object, we apply the following recovery mechanism. The system goes back to the detection stage, with a query feature descriptor at hand as \( v = \frac{1}{|V_{obj}|} \sum_{v_{obj} \in V_{obj}} v_{obj} \), seeking the closest region from the segmented frame, and thus re-detecting the object. Here, the segmentation might be given by the segmentation model (e.g., SAM), or by the tracker which tries to re-detect the lost object. The means to gain faster performance for real-time applications.

\begin{figure*}[!htbp]
    \centering
    \includegraphics[width=0.6\textwidth]{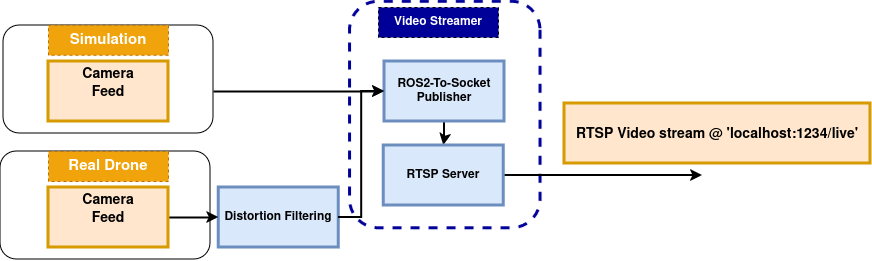}
    \caption{Redesigned input pipeline for publishing the camera feed as an RTSP video stream.}
    \label{fig:input_source}
\end{figure*}

\begin{figure*}[!htbp]
    \centering
    \includegraphics[width=0.8\textwidth]{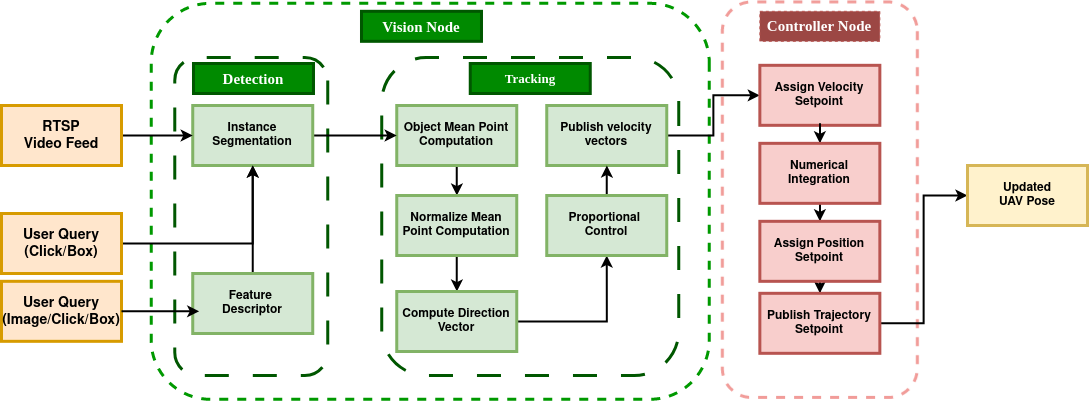}
    \caption{Process Flow of the Project TAR}
    \label{fig:process_flow}
\end{figure*}

\section{Experiment}
\subsection{Hardware}
Our project utilizes a MODALAI M500 drone equipped with a VOXL2 flight deck, incorporating the Qualcomm Flight RB5 5G Platform and QRB5165 processor. This setup is integrated with PX4 on DSP for flight control and performance. The drone also equips a black and white 640x480 tracking camera that streams live footage to the ground control station (GCS) using RTSP. Control commands are managed through a FrSky Taranis Q X7 radio controller with an R9M 900MHz transmitter, ensuring reliable communication. The built drone used for the experiments can be seen in \ref{fig:rapter}.

To track the ground truth of the Rapter and the target, we have installed five Vicon markers on the drone, enabling precise location data. The GCS is powered by a computer equipped with an NVIDIA GeForce 4060 graphics card and an Intel Core i7 13620H processor, providing the necessary computational power to handle complex processing tasks. The system operates using ROS2, which enables efficient communication and data handling between the drone and the GCS. This setup ensures real-time tracking, robust control, and high-performance operation of the drone, suitable for various advanced applications.

\subsection{Simulation}
\begin{figure*}[!htbp]
    \centering
    \includegraphics[width=0.8\textwidth]{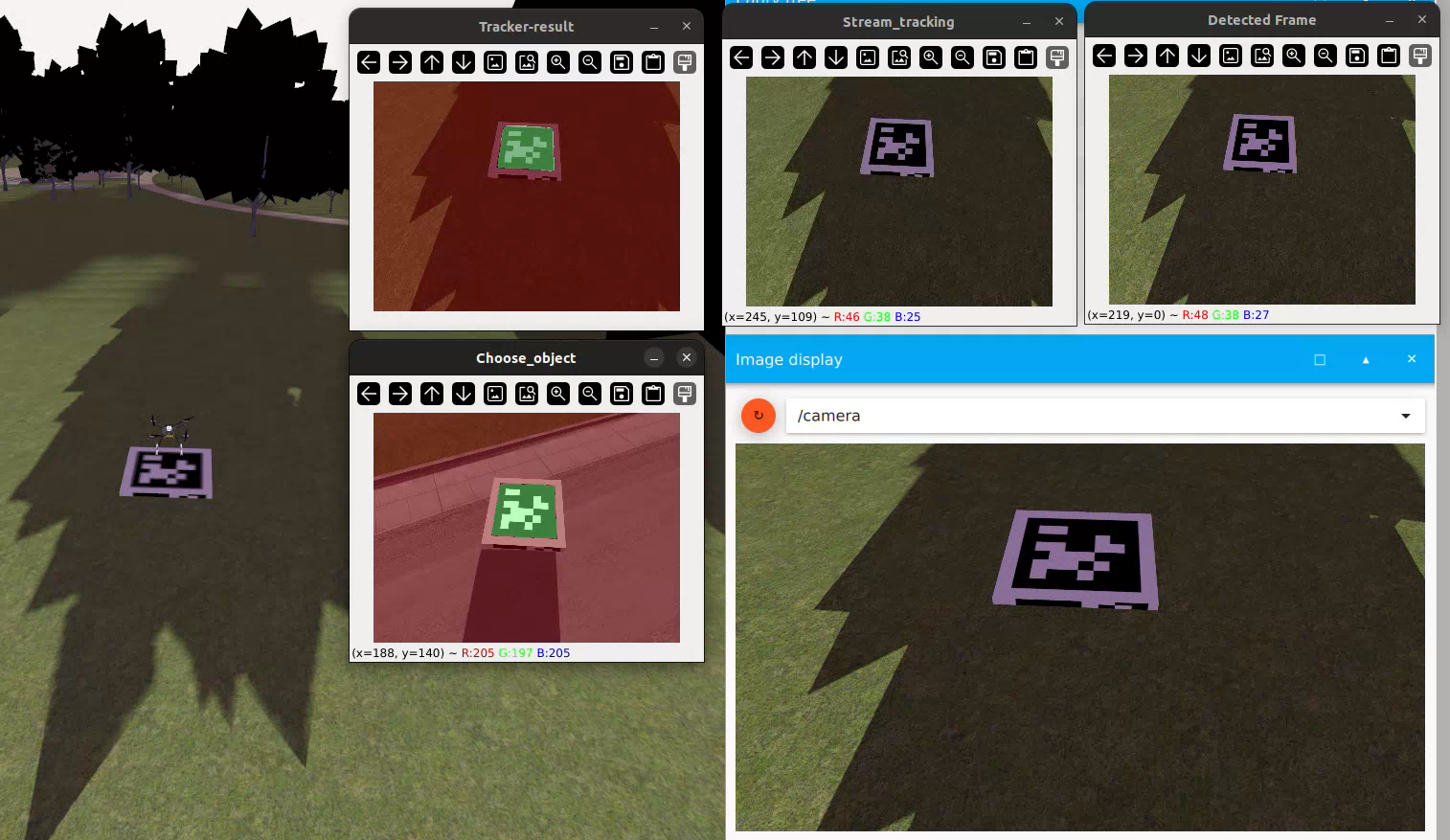}
    \caption{Oneshot Apriltag tracking based on the real-time bounding box query.}
    \label{fig:sim}
\end{figure*}

The implemented tracking algorithm was evaluated in a ROS2 Gazebo environment, where a PX4-based simulator drone was tasked with tracking an Apriltag in motion. The tracking was performed on the fly, based on a bounding box query provided around the Apriltag model, without any prior training. The results can be seen in \ref{fig:sim}.

\subsection{Implementation Details}
The key details involved in the implementation of the project are as follows: 
\subsubsection{Input Source}
The primary input source is the real-time camera feed from a real Unmanned Aerial Vehicle (UAV) or a simulated UAV. As a design choice, these feeds are converted into Real-Time Streaming Protocol (RTSP) video streams. This conversion allows the feeds to be accessed by any device within the same network. This design facilitates performing all computationally intensive model inferences externally, rather than on the drone itself. During the mission, only the control signals, generated based on the processing of the real-time feed, are sent to the inbuilt low-level controller within the PX4 Autopilot on the drone.

A ROS2-based node is deployed to subscribe to the 'camera' topic, obtaining the real-time camera feed from the real UAV (filtered to avoid distortion) and the simulation UAV. The obtained frames from the camera feed are converted from ROS format to OpenCV format. These frames are then resized, encoded to JPEG format, and subsequently converted to a base64 string. The base64 encoded string is transmitted over the network via the socket protocol to port '5555'. An RTSP server using GStreamer is employed to decode, process, and stream the real-time camera feed as an RTSP video feed. The video feed is accessible at the URL 'localhost:1234/live', as illustrated in the accompanying figure \ref{fig:input_source}. This approach ensures efficient handling of real-time video streams, enabling robust external processing while maintaining effective control signal transmission to the UAV's low-level controller.

\subsubsection{Process Flow}
In this system, the drone's movement is adjusted dynamically based on the position of an object within the video frame. 

The processing of the incoming video feed and the generation of appropriate control signals are achieved using two ROS2 nodes as depicted in figure \ref{fig:process_flow}. The first node, named the Vision node, is responsible for processing the incoming visual feed to generate the appropriate control signals. The second node, named the Controller node, is responsible for utilizing the low-level controller to execute motions based on the obtained control signals.

The Vision node is designed to control the UAV drone's movement based on the position of a tracked object within a video frame. The core principle involves calculating a direction vector, 
\[
\mathbf{d} = \left[\frac{x}{W} - 0.5, -\left(\frac{y}{H} - 0.5\right), \theta\right],
\]
where \( x \) and \( y \) represent the coordinates of the object's mean point, \( W \) and \( H \) are the frame's width and height, respectively, and \( \theta \) is the heading angle derived from the arctangent function,
\[
\theta = -\arctan\left(\frac{\Delta x}{\Delta y}\right).
\]

The direction vector is subsequently scaled by gain factors \( K_X \), \( K_Y \), \( K_Z \), and \(\text{yaw}_K\), yielding the control signals
\[
\mathbf{u} = [K_X d_x, K_Y d_y, K_Z d_z, \text{yaw}_K \theta],
\]
which determines the drone's velocity and heading adjustments. To enhance robustness, a filtering process utilizing the coefficients \(\mathbf{b}\) and \(\mathbf{a}\) is applied to smooth the control signals, mitigating the effects of noise and abrupt positional changes. This results in a filtered direction vector, 
\[
\mathbf{d}_{\text{filt}}.
\]

In scenarios where the object is lost, the function clears the positional queue and issues a reset signal to the ROS node, enabling the drone to re-establish tracking. The use of yaw control ensures accurate heading adjustments, calculated by
\[
\theta = -\arctan\left(\frac{\Delta x}{\Delta y}\right) \cdot \frac{180}{\pi},
\]
to align the drone with the target object. This systematic integration of real-time object detection, proportional control, and filtering facilitates precise and stable drone maneuvering, ensuring effective object tracking within dynamic environments.

The Controller node is a ROS2 node that controls the UAV in offboard mode using the PX4 Autopilot. The node subscribes to various topics to receive real-time data about the UAV's status and position, processes incoming target pose data, and publishes control signals to guide the UAV's movements.

The node operates by maintaining a continuous control loop, in which it periodically publishes offboard control mode signals to ensure the UAV remains in offboard mode. The core functionality is triggered upon receiving target pose data from the \textit{tracked\_pose} topic, which contains the desired velocity and yaw rate for the UAV. Upon receiving this data, the node calculates the new positional setpoints by integrating the target velocities over a discrete time step \(\Delta t\), updating the UAV's current position \(\mathbf{p}_{\text{current}}\) to the new target position \(\mathbf{p}_{\text{target}}\):

\[
\mathbf{p}_{\text{target}}(t) = \mathbf{p}_{\text{current}} + \mathbf{v}_{\text{target}} \cdot \Delta t
\]

where \(\mathbf{v}_{\text{target}} = [v_x, v_y, v_z]\) represents the target velocity components along the x, y, and z axes. Additionally, the yaw angle \(\psi\) is updated based on the target yaw rate \(\dot{\psi}_{\text{target}}\):

\[
\psi_{\text{target}} = \psi_{\text{current}} + \dot{\psi}_{\text{target}} \cdot \Delta t
\]

The node then creates a `TrajectorySetpoint` message containing these updated position, velocity, and yaw setpoints and publishes it to guide the UAV's trajectory. This ensures that the UAV follows the desired path with the specified velocity and orientation. The node continuously monitors the UAV's local position and status to provide real-time feedback, enhancing control accuracy and stability.

\begin{figure}[!htbp]
    \centering
    \begin{subfigure}[b]{0.48\columnwidth}
        \centering
        \includegraphics[width=\textwidth]{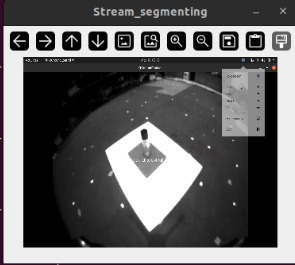}
        \caption{Streamed input to the Tracking model}
        \label{fig:stream_input}
    \end{subfigure}
    \hfill
    \begin{subfigure}[b]{0.48\columnwidth}
        \centering
        \includegraphics[width=\textwidth]{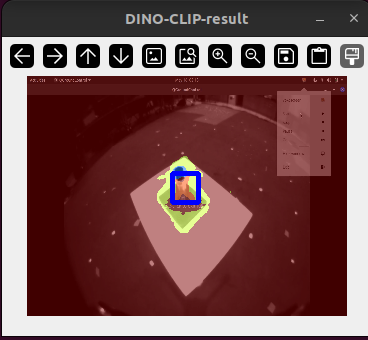}
        \caption{DINO-based feature matching on streamed input}
        \label{fig:dino_feature_extraction}
    \end{subfigure}
    \vskip\baselineskip
    \begin{subfigure}[b]{0.48\columnwidth}
        \centering
        \includegraphics[width=\textwidth]{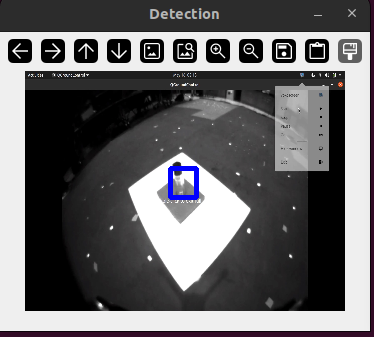}
        \caption{Bounding Box extraction from the DINO discovered pixel map.}
        \label{fig:dino_detection}
    \end{subfigure}
    \hfill
    \begin{subfigure}[b]{0.48\columnwidth}
        \centering
        \includegraphics[width=\textwidth]{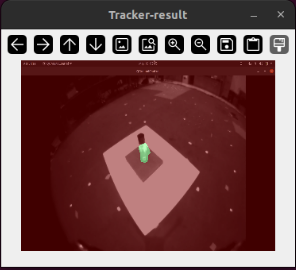}
        \caption{Tracking the DINO generated pixelmap using AOT}
        \label{fig:image4}
    \end{subfigure}
    \caption{Image template based One-shot tracking pipeline using DINO}
    \label{fig:Dino_pipeline}
\end{figure}

\begin{figure}[!htbp]
    \centering
    \begin{subfigure}[b]{0.48\columnwidth}
        \centering
        \includegraphics[width=\textwidth]{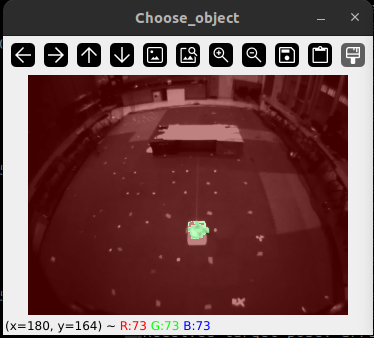}
        \caption{Chosen input via bounding box}
        \label{fig:bbox_stream_input}
    \end{subfigure}

    \vskip\baselineskip
    \begin{subfigure}[b]{0.48\columnwidth}
        \centering
        \includegraphics[width=\textwidth]{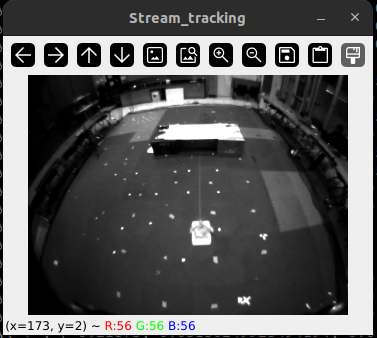}
        \caption{Streamed video feed while tracking is active}
        \label{fig:bbox_active_stream}
    \end{subfigure}
    \hfill
    \begin{subfigure}[b]{0.48\columnwidth}
        \centering
        \includegraphics[width=\textwidth]{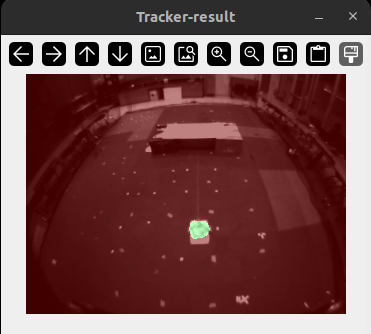}
        \caption{Tracking the generated pixelmap using AOT}
        \label{fig:bbox_tracking}
    \end{subfigure}
    \caption{Bounding Box-based One-shot tracking pipeline}
    \label{fig:bbox_pipeline}
\end{figure}

\section{Results}
The repository containing the code for the work can be found \href{https://github.com/tvpian/Project-TAR}{here}.

Demonstration videos for all the experiments and the note describing what each video depicts can be found \href{https://drive.google.com/drive/u/1/folders/1gO0R1qUqjNkcfFNRyeN-Uf0ZnvGIiMT8} {here}. 

\subsection{Evaluation Metric}
Dynamic Time Warping (DTW) is utilized to compare two sequences of positions: $\text{positions1}$ representing the trajectory of the drone and $\text{positions2}$ denoting the trajectory of the target. These sequences, $\text{positions1} = \{ \mathbf{p}_1^1, \mathbf{p}_2^1, \ldots, \mathbf{p}_n^1 \}$ and $\text{positions2} = \{ \mathbf{p}_1^2, \mathbf{p}_2^2, \ldots, \mathbf{p}_m^2 \}$, consist of position vectors $\mathbf{p}_i^1$ and $\mathbf{p}_j^2$ representing the respective positions of the drone and target at time steps $i$ and $j$.

The distance and path between the positions can be computed using the FastDTW algorithm: \(\text{distance}, \text{path} = \text{fastdtw}(\text{positions1}, \text{positions2}, \text{dist}(\mathbf{x}, \mathbf{y}) = \left\| \mathbf{x} - \mathbf{y} \right\|)\)

Here, $\text{fastdtw}$ calculates the DTW distance between the sequences, incorporating both temporal and spatial variations. The resulting $\text{distance}$ represents the cumulative distance between corresponding points of the two trajectories, while $\text{path}$ indicates the optimal alignment between the sequences.

The distance and path between the positions can be computed using the FastDTW algorithm:
\begin{align*}
\text{distance}, \text{path} &= \text{fastdtw}(\text{positions1}, \text{positions2},\\
& \quad \text{dist}(\mathbf{x}, \mathbf{y}) = \left\| \mathbf{x} - \mathbf{y} \right\|)
\end{align*}

Furthermore, the algorithm determines the individual distances \(\text{dtw\_distances}\) for each matched pair of positions along the DTW path:
\[
\text{dtw\_distances} = \left\| \mathbf{p}_i^1 - \mathbf{p}_j^2 \right\| \quad \text{for} \quad (i, j) \in \text{path}
\]

These distances quantify the dissimilarity between the drone and target trajectories at specific time instances. Finally, the mean distance $\text{mean\_distance}$ is computed as the average of $\text{dtw\_distances}$, offering a comprehensive measure of the overall dissimilarity between the trajectories.

A total of eight runs were conducted, utilizing various targets and modalities. The resulting trajectory paths were plotted, and Dynamic Time Warping (DTW) was applied to compare the drone's path to the target's path. This analysis allowed us to evaluate the accuracy of the drone's tracking performance across different modalities and targets. By calculating the mean DTW values, we determined that the 'Dino' feature extractor yielded the lowest mean value, indicating superior tracking accuracy compared to the other modalities.

\begin{table}[!h]
\begin{tabular}{cclll}
\textbf{Case}           & \textbf{Mean DTW(m)} & {\color[HTML]{000000} } &  &  \\
turtle\_fan\_bb\_no     & 0.98                 & {\color[HTML]{000000} } &  &  \\
turtle\_fan\_dino\_no   & 0.62                 & {\color[HTML]{000000} } &  &  \\
apriltag\_fan\_bb\_no   & 0.98                 & {\color[HTML]{000000} } &  &  \\
apriltag\_fan\_dino\_no & 0.67                 & {\color[HTML]{000000} } &  &  \\
apriltag\_fan\_dino\_o  & 1.56                 & {\color[HTML]{000000} } &  &  \\
turtle\_fan\_dino\_o    & 1.65                 & {\color[HTML]{000000} } &  &  \\
apriltag\_vtm\_node     & 0.8                  & {\color[HTML]{000000} } &  & 
\end{tabular}
\caption{Mean DTW values for different cases}
\label{tab:mean-dtw}
\end{table}

The naming convention used in this study to represent each tracking case follows the format \textit{target\_algorithm\_modality\_obstruction}. The \textit{target} refers to the object being tracked, either \textit{turtle} or \textit{apriltag}. The \textit{algorithm} denotes the tracking method employed, such as \textit{fan} for the "Follow Anything" algorithm or \textit{vtm} for the \textit{voxl\_to\_mpa\_ros2} algorithm. The \textit{modality} indicates the type of data used for tracking, with \textit{bb} representing the bounding box and \textit{DINO} referring to the dino modality. Lastly, the \textit{obstruction} specifies whether there is any obstruction in the tracking environment, denoted as \textit{o} for obstruction present or \textit{no} for no obstruction. For instance, \textit{turtle\_fan\_bb\_no} refers to tracking a turtle target using the "Follow Anything" algorithm with bounding box modality in an environment without obstructions.

\begin{figure}[!htpb]
     \centering
    \includegraphics[width=\linewidth]{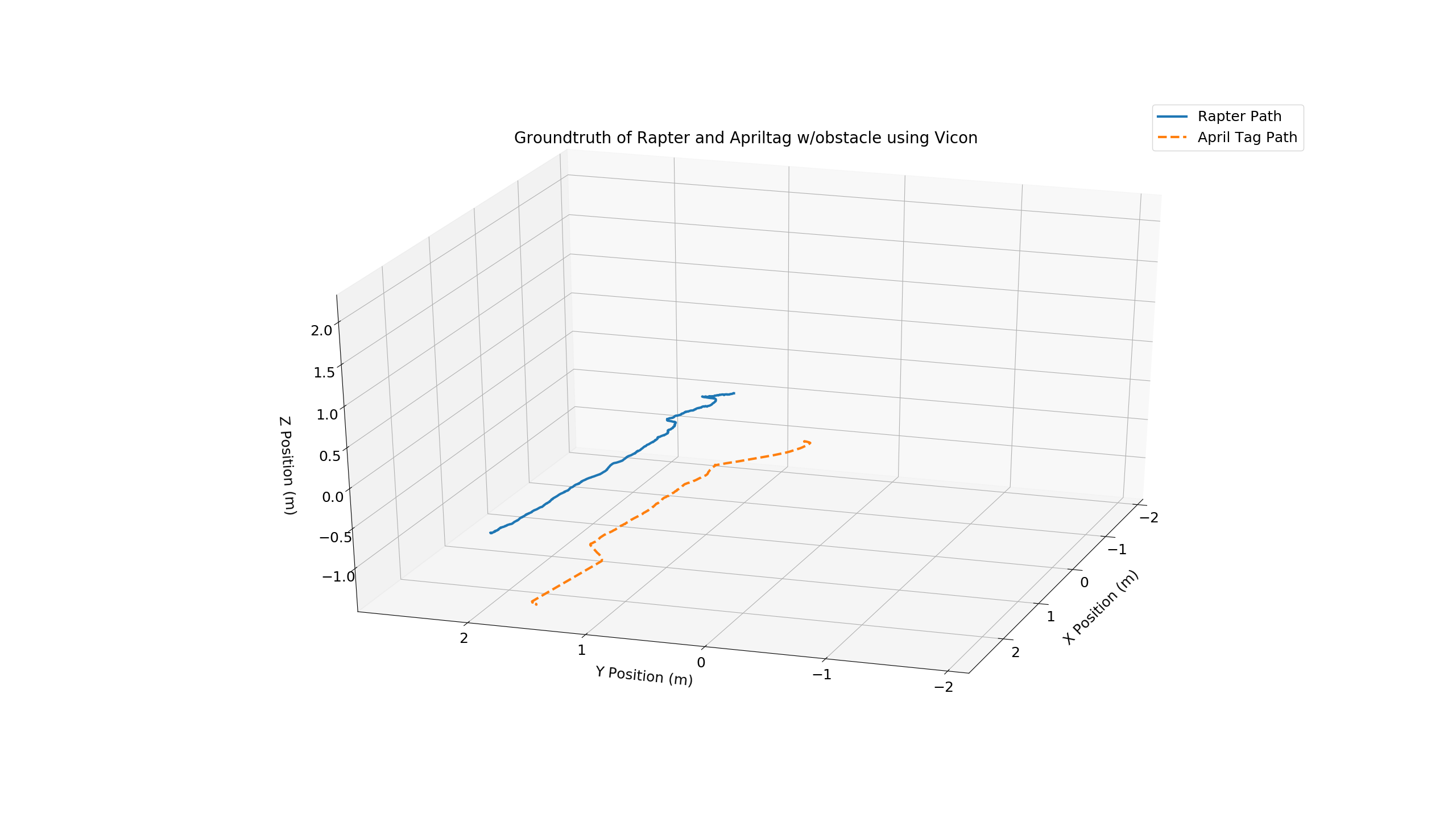}
    \caption{Rapter Trajectory with Apriltag Trajectory using Apriltag Tracking with voxl mpa to ros2 method}
    \label{fig:Apriltag tracking w/voxl_to_mpa_ros2 method}

    \includegraphics[width=0.75\linewidth]{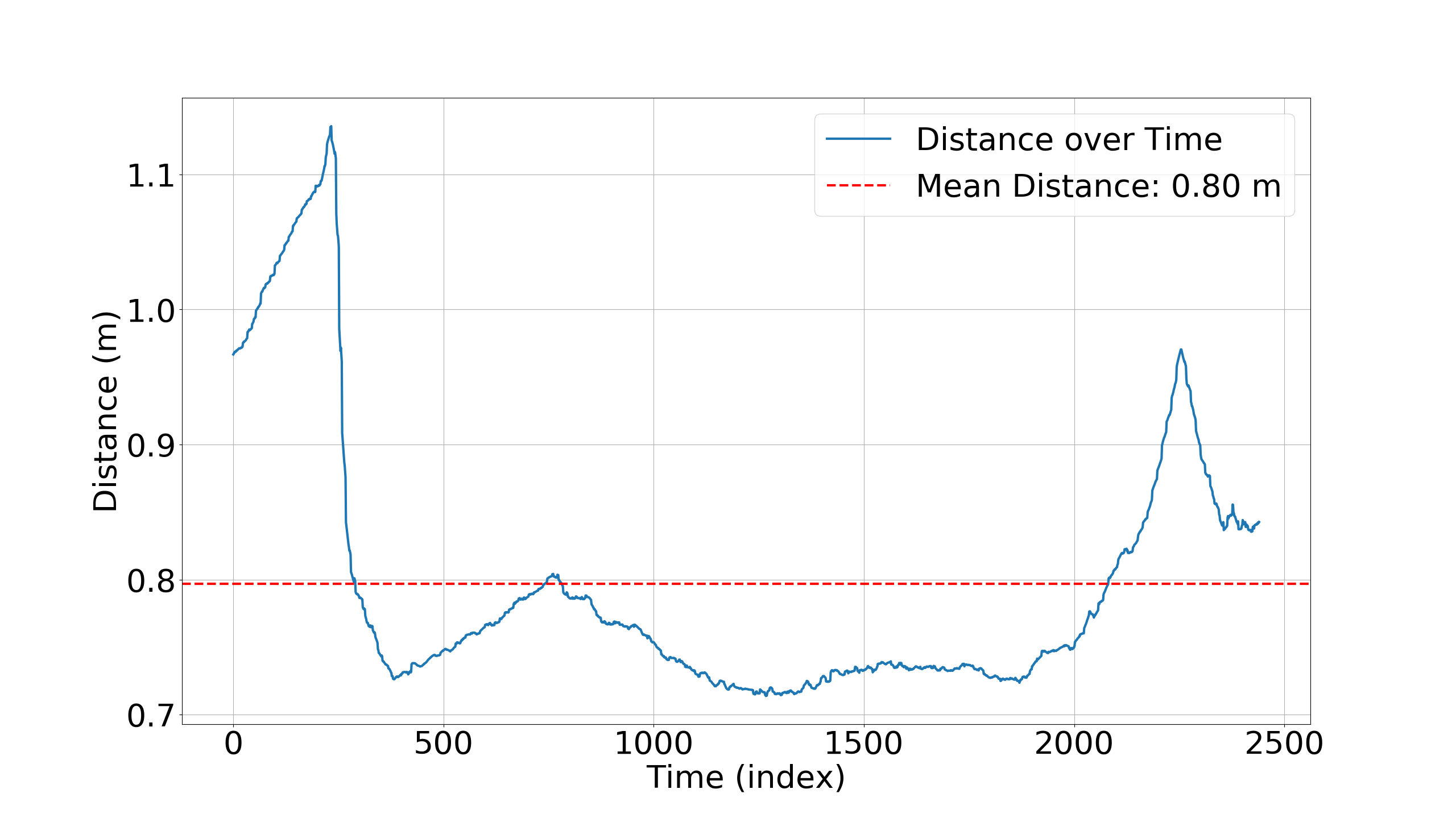}
    \caption{DTW(Dynamic Time Warping) of Rapter Trajectory with Apriltag Trajectory using Apriltag Tracking with voxl mpa to ros2 method}
    \label{fig:Apriltag tracking w/voxl_to_mpa_ros2 method dtw}
    
    
\end{figure}

\begin{figure}[!htpb]
     \centering
    \includegraphics[width=\linewidth]{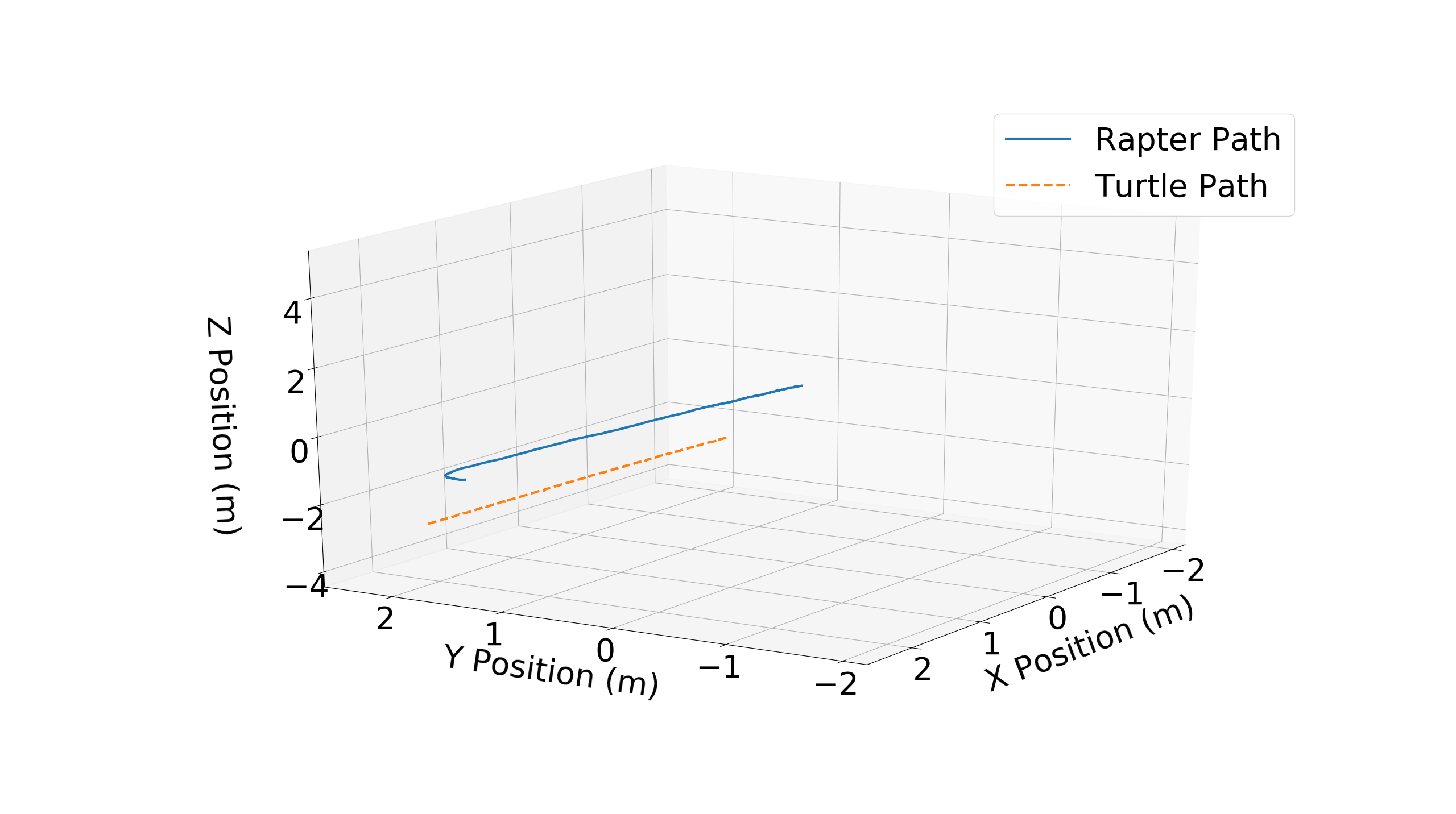}
    \caption{Rapter Trajectory with Apriltag Trajectory}
    \label{fig:apriltag_fan_no}

    \includegraphics[width=0.75\linewidth]{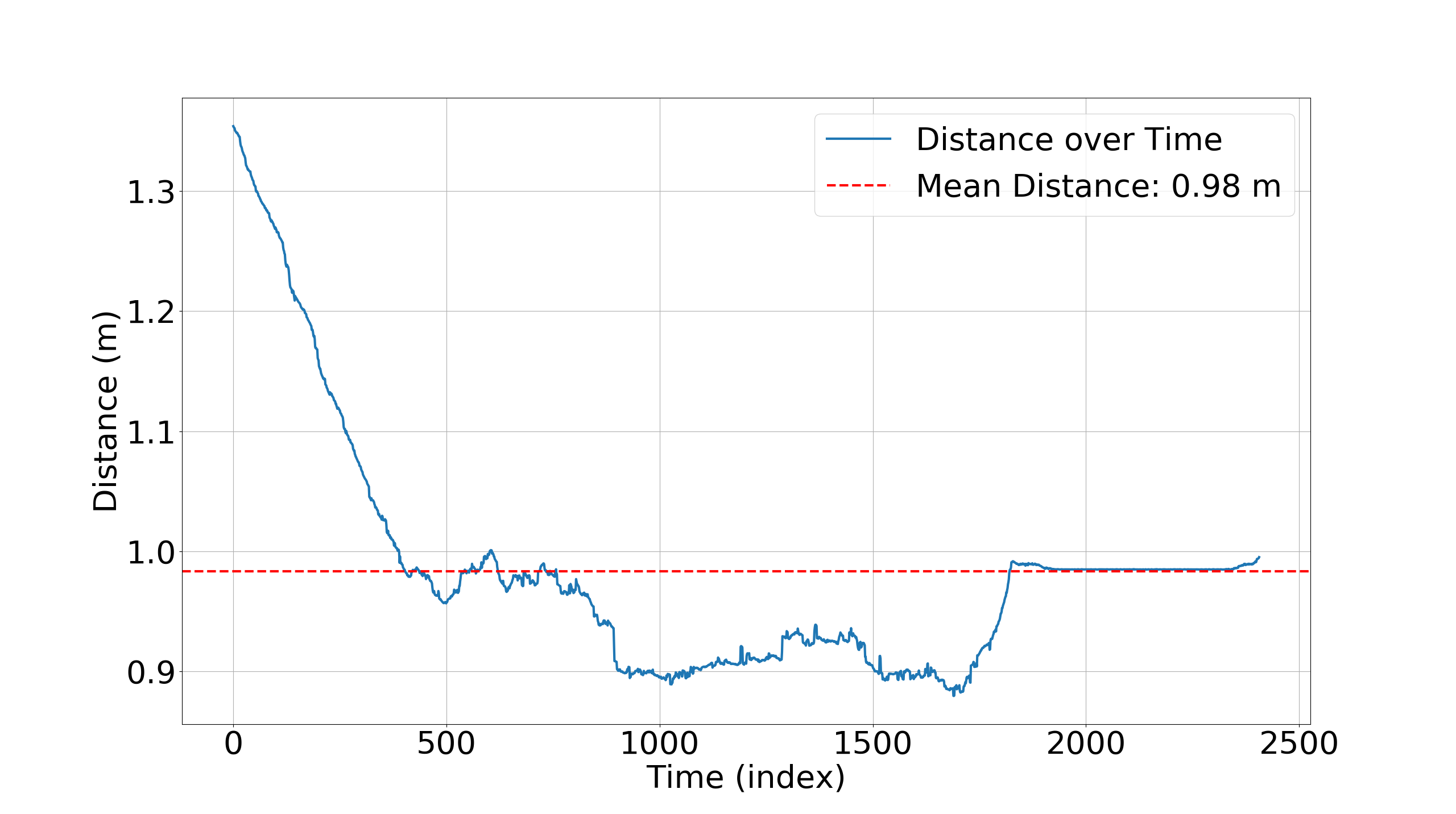}
    \caption{DTW(Dynamic Time Warping) of Rapter Trajectory with Apriltag Trajectory}
    \label{fig:apriltag_fan_no_dtw}    
\end{figure}

\begin{figure}[!htpb]
     \centering
    \includegraphics[width=\linewidth]{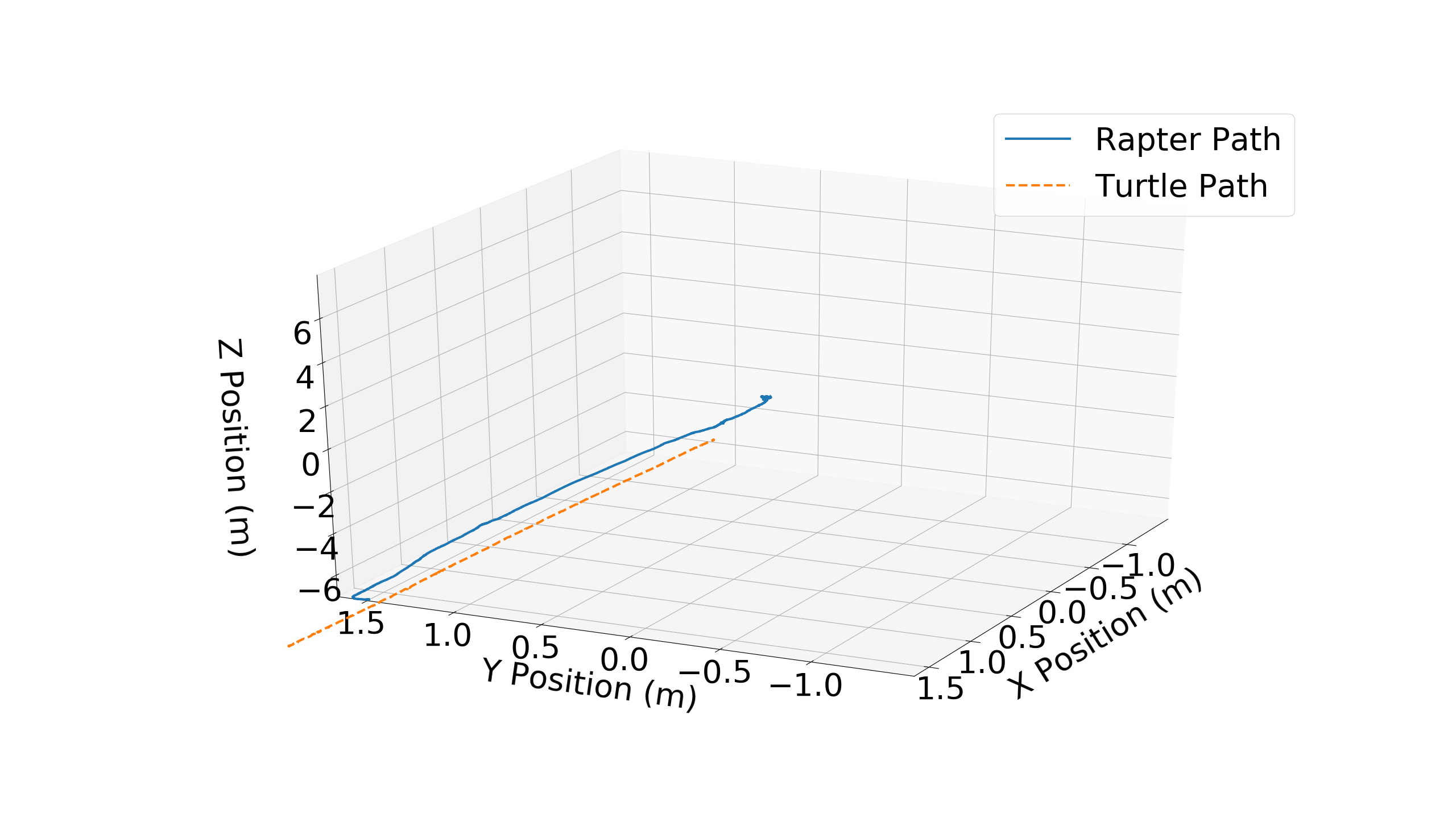}
    \caption{Rapter Trajectory with Turtle Trajectory}
    \label{fig:turtle_fan_no}

    \includegraphics[width=0.75\linewidth]{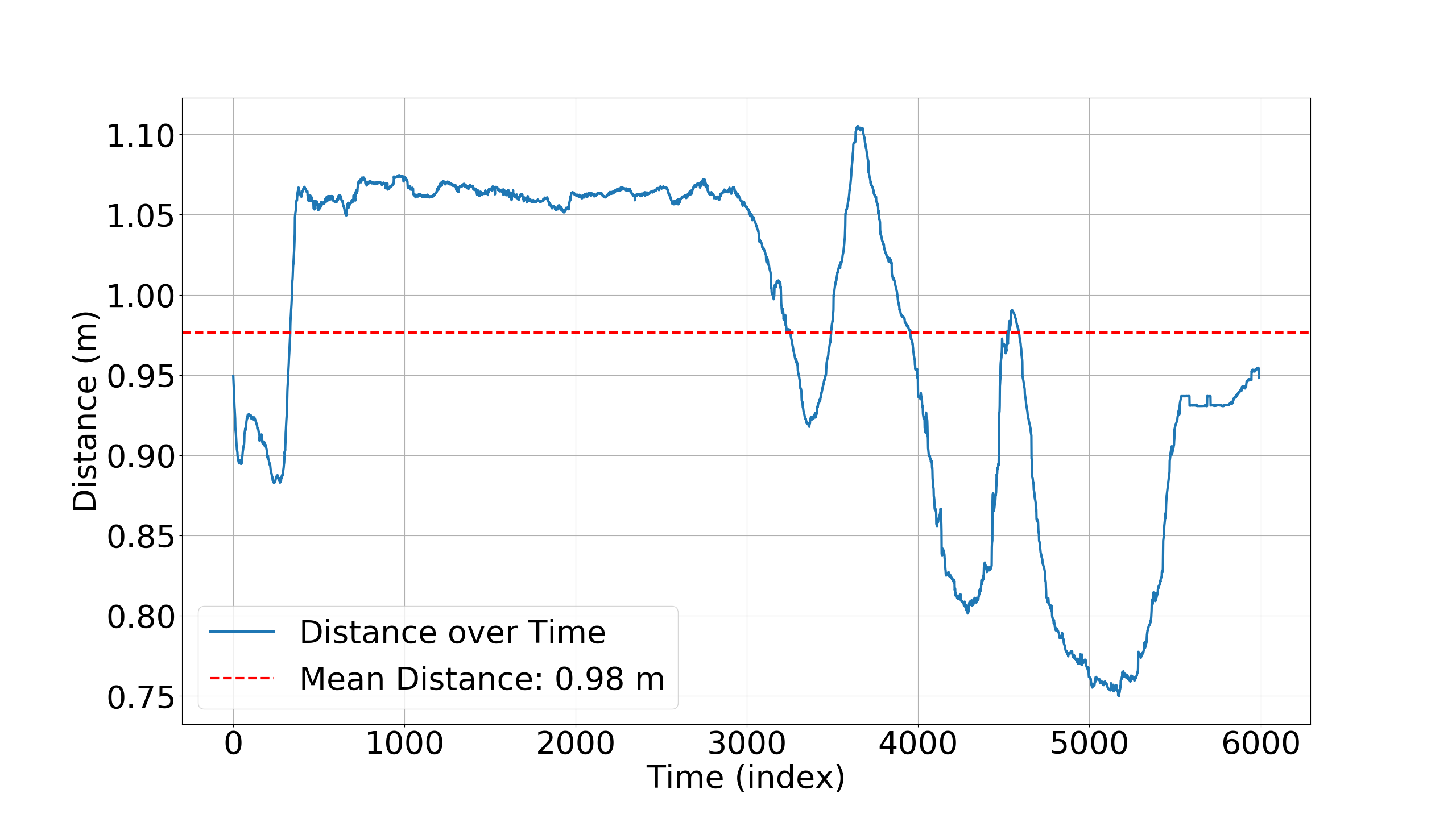}
    \caption{DTW(Dynamic Time Warping) of Rapter Trajectory with Turtle Trajectory}
    \label{fig:turtle_fan_no_dtw}
    
    
\end{figure}

\begin{figure}[!htpb]
     \centering
    \includegraphics[width=\linewidth]{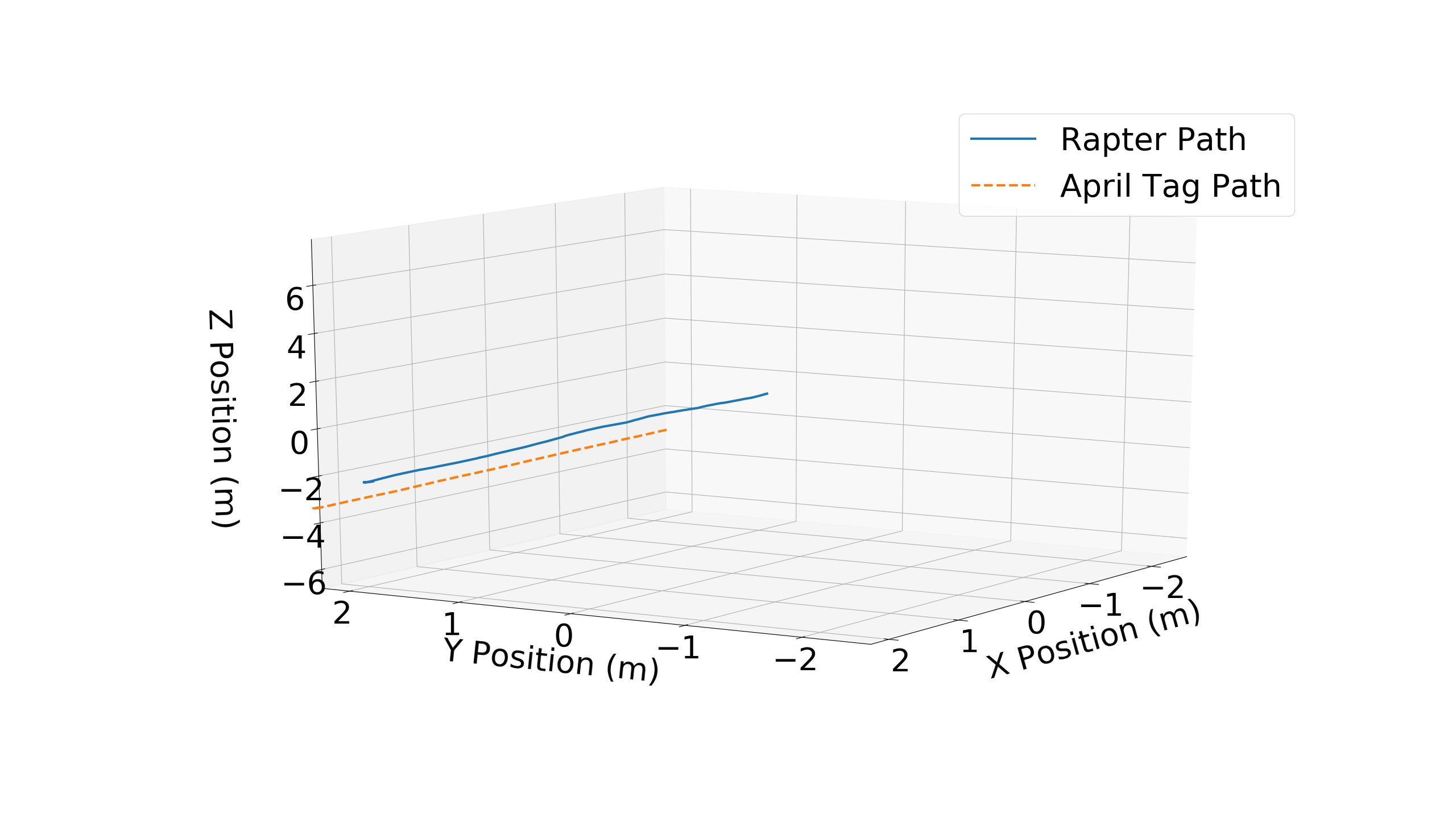}
    \caption{Rapter Trajectory with Apriltag Trajectory using DINO}
    \label{fig:apriltag_fan_dino_no}

    \includegraphics[width=0.75\linewidth]{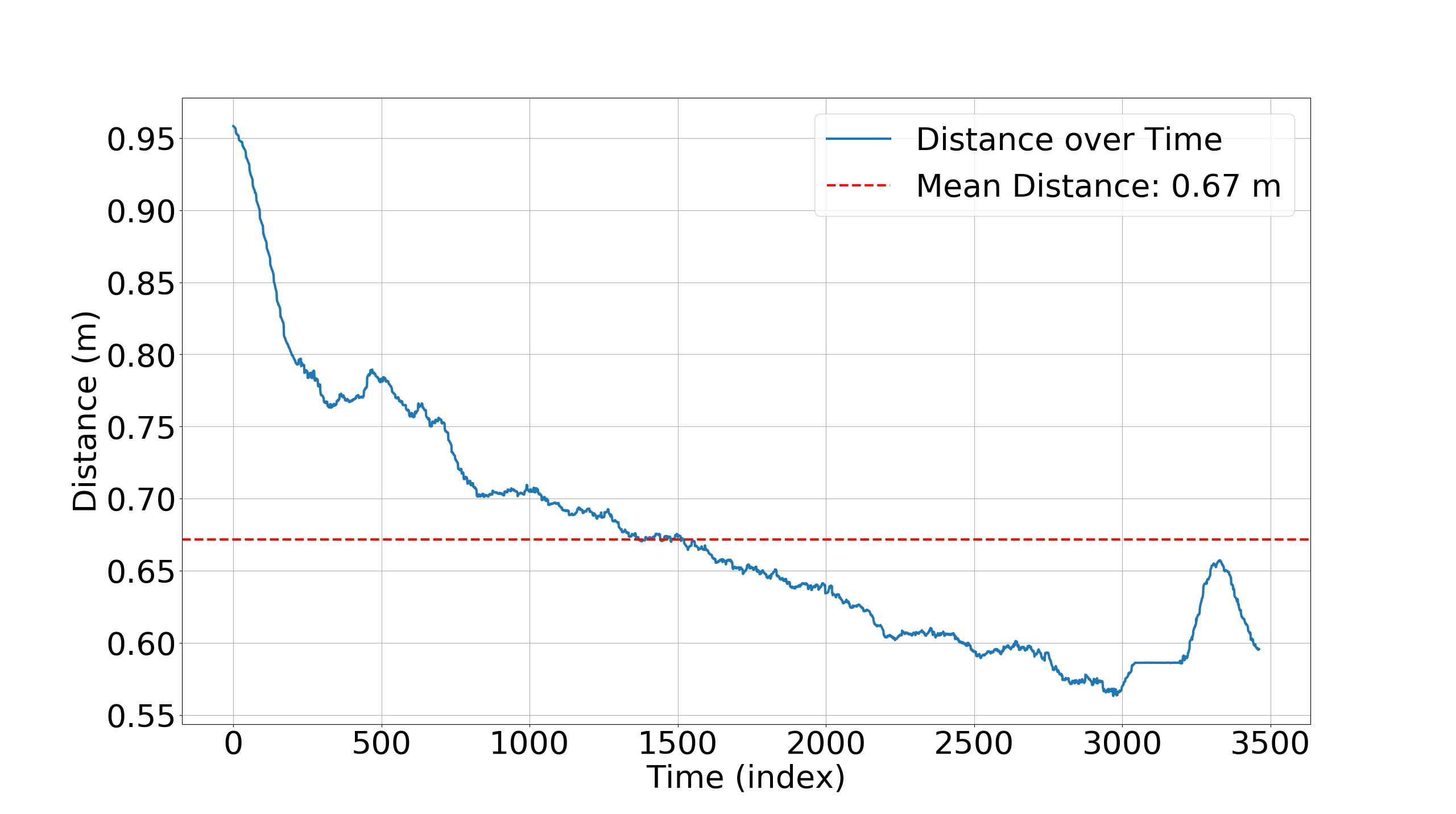}
    \caption{DTW(Dynamic Time Warping) of Rapter Trajectory with Apriltag Trajectory using DINO}
    \label{fig:apriltag_fan_dino_no_dtw}
\end{figure}

\begin{figure}[!htpb]
     \centering
    \includegraphics[width=\linewidth]{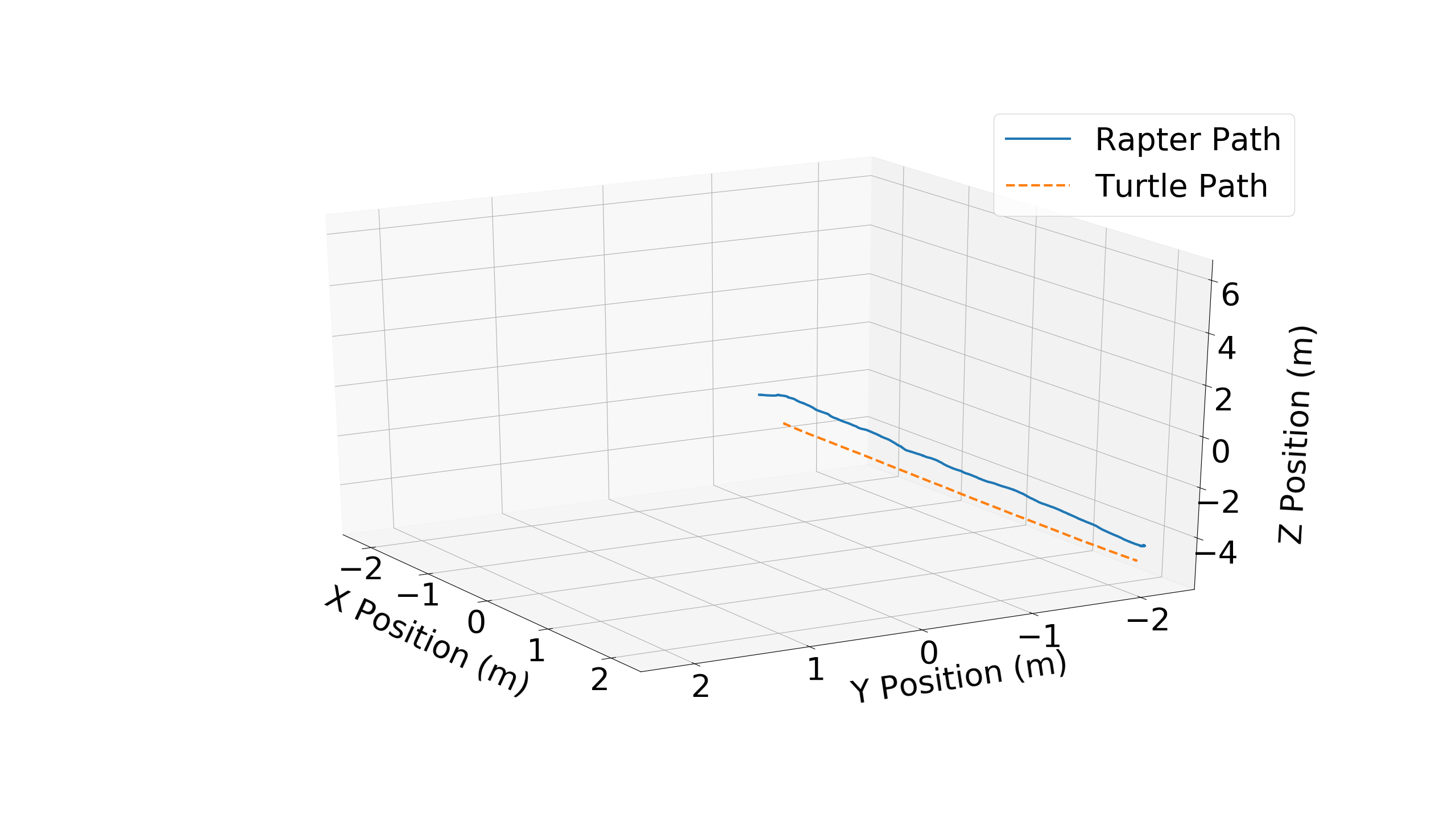}
    \caption{Rapter Trajectory with Turtle Trajectory using DINO}
    \label{fig:turtle_fan_dino_no}

    \includegraphics[width=0.75\linewidth]{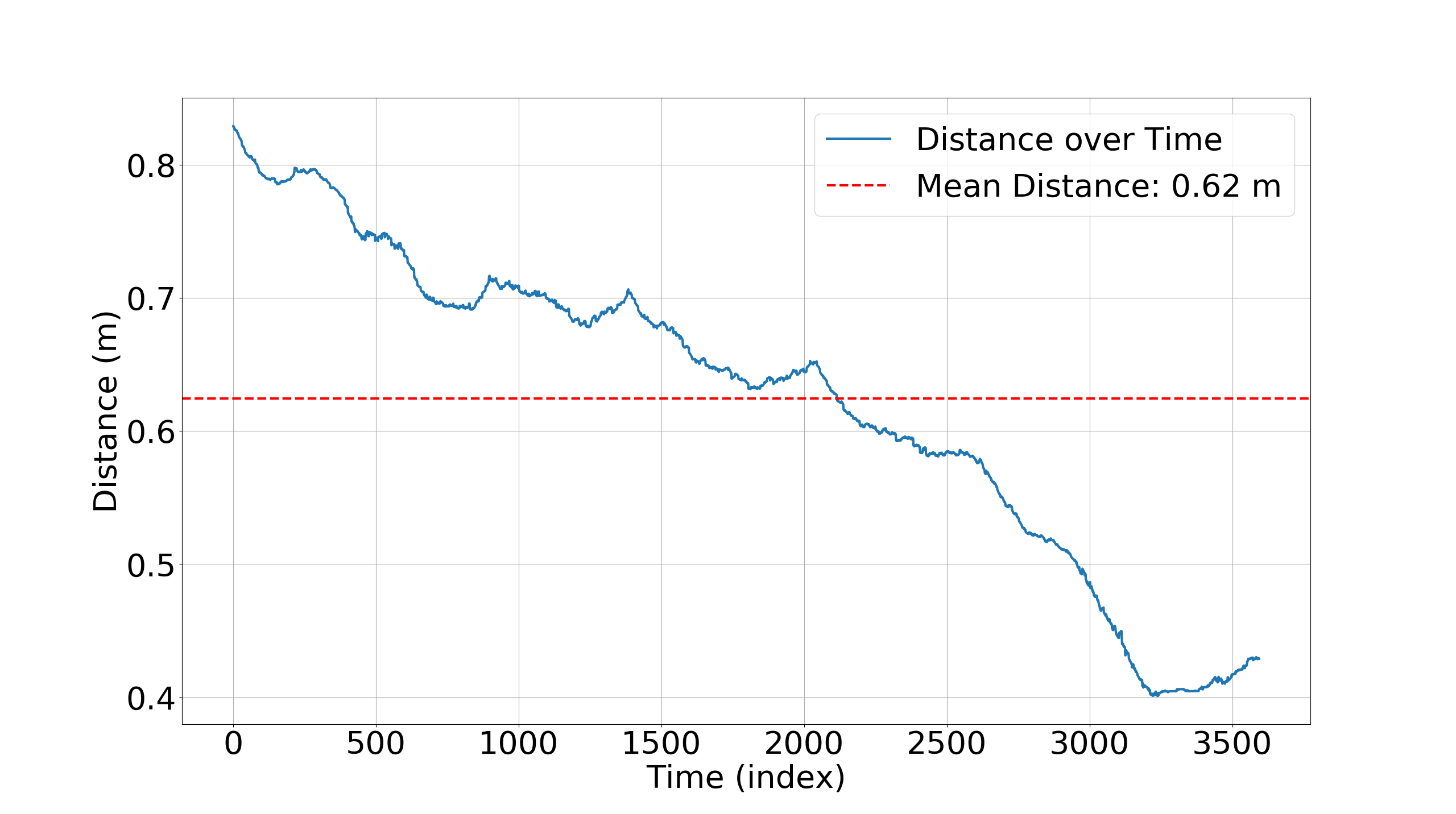}
    \caption{DTW(Dynamic Time Warping) of Rapter Trajectory with Turtle Trajectory using DINO}
    \label{fig:turtle_fan_dino_no_dtw}
    
    
\end{figure}

\begin{figure}[!htpb]
     \centering
    \includegraphics[width=\linewidth]{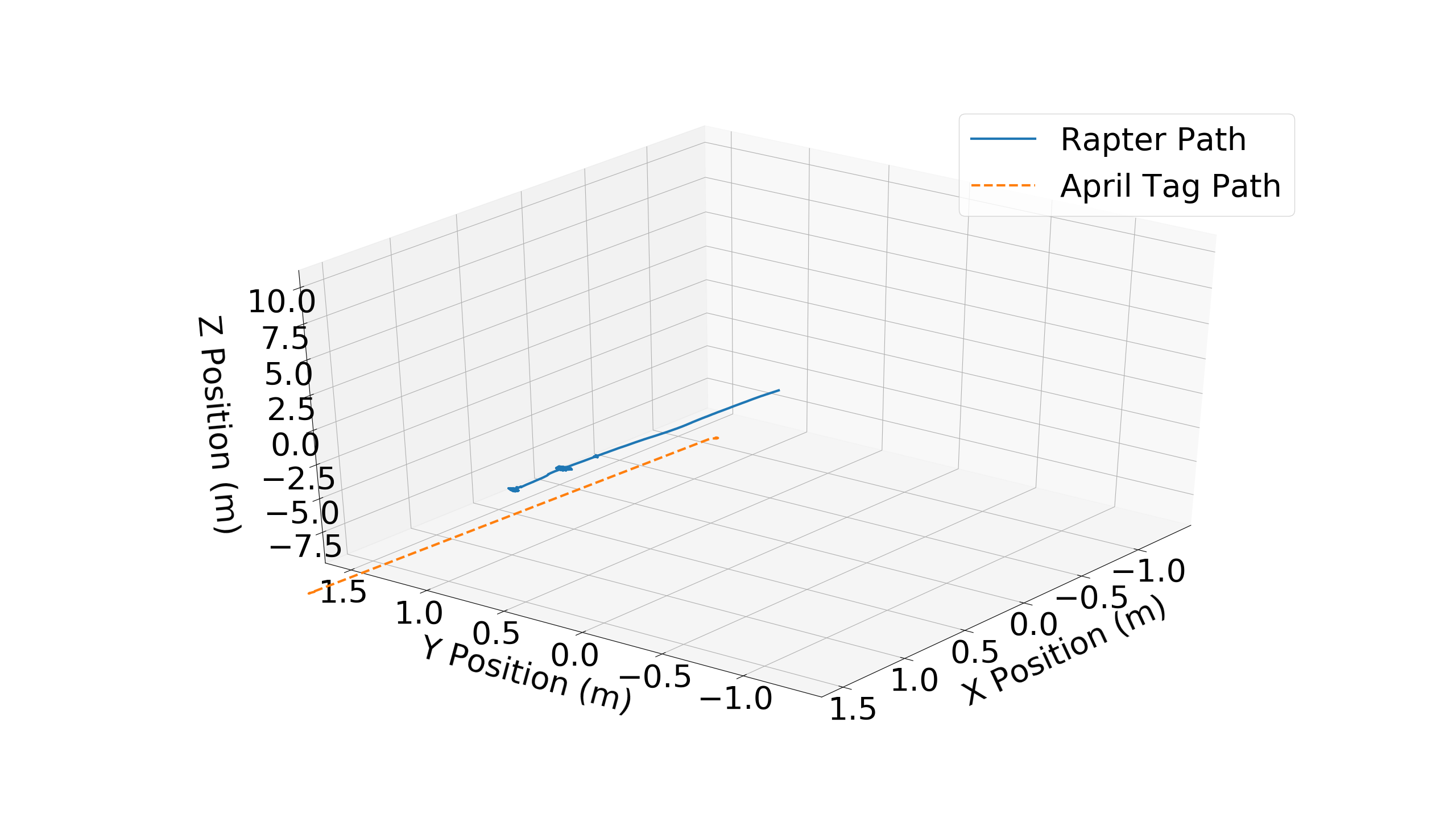}
    \caption{Rapter Trajectory with Apriltag Trajectory using DINO w/obstruction}
    \label{fig:apriltag_fan_dino_o}

        \includegraphics[width=0.75\linewidth]{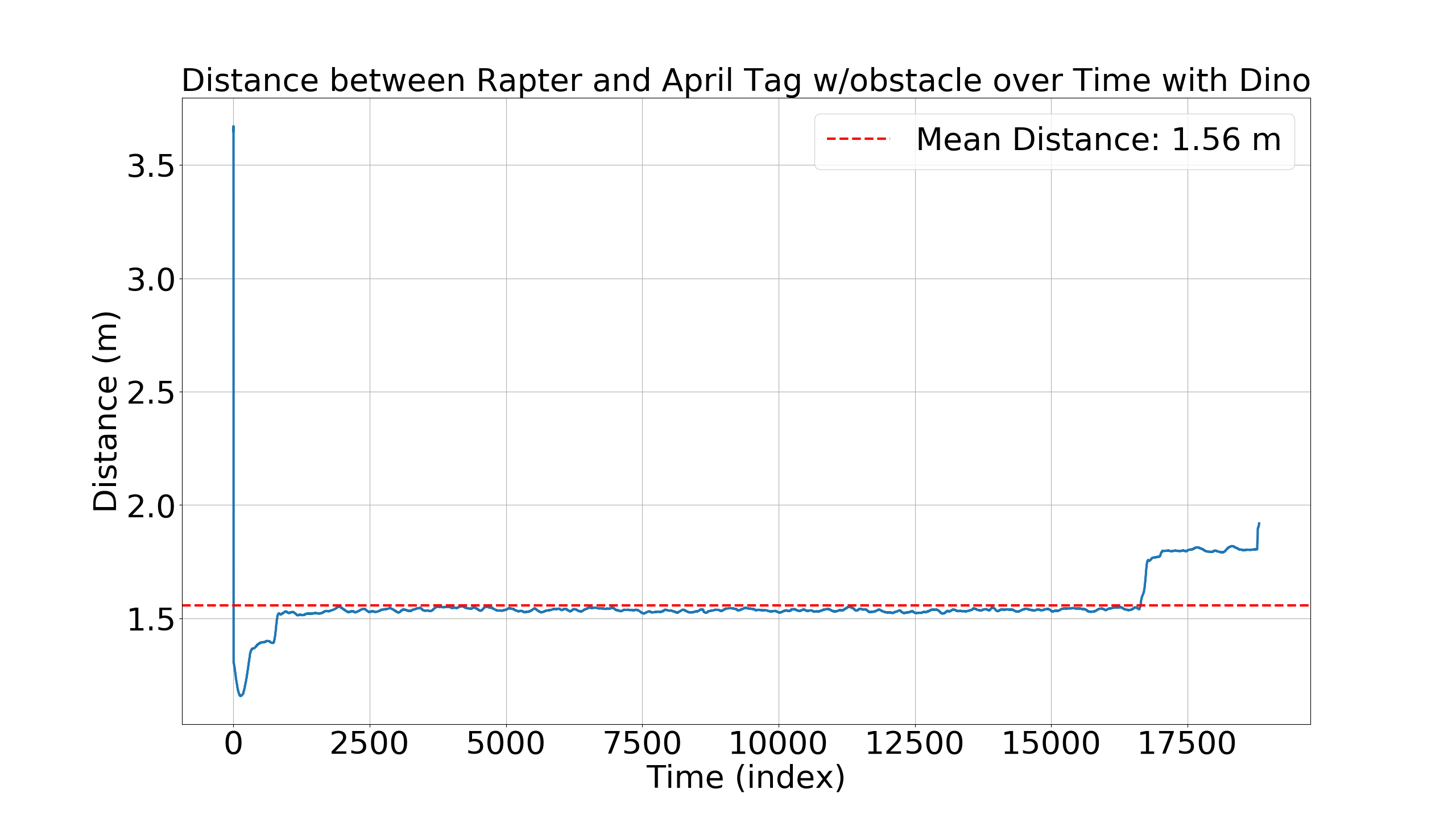}
    \caption{DTW(Dynamic Time Warping) of Rapter Trajectory with Apriltag Trajectory using DINO w/obstruction}
    \label{fig:apriltag_fan_dino_o_dtw}
    
    
\end{figure}

\begin{figure}[!htbp]
     \centering
    \includegraphics[width=\linewidth]{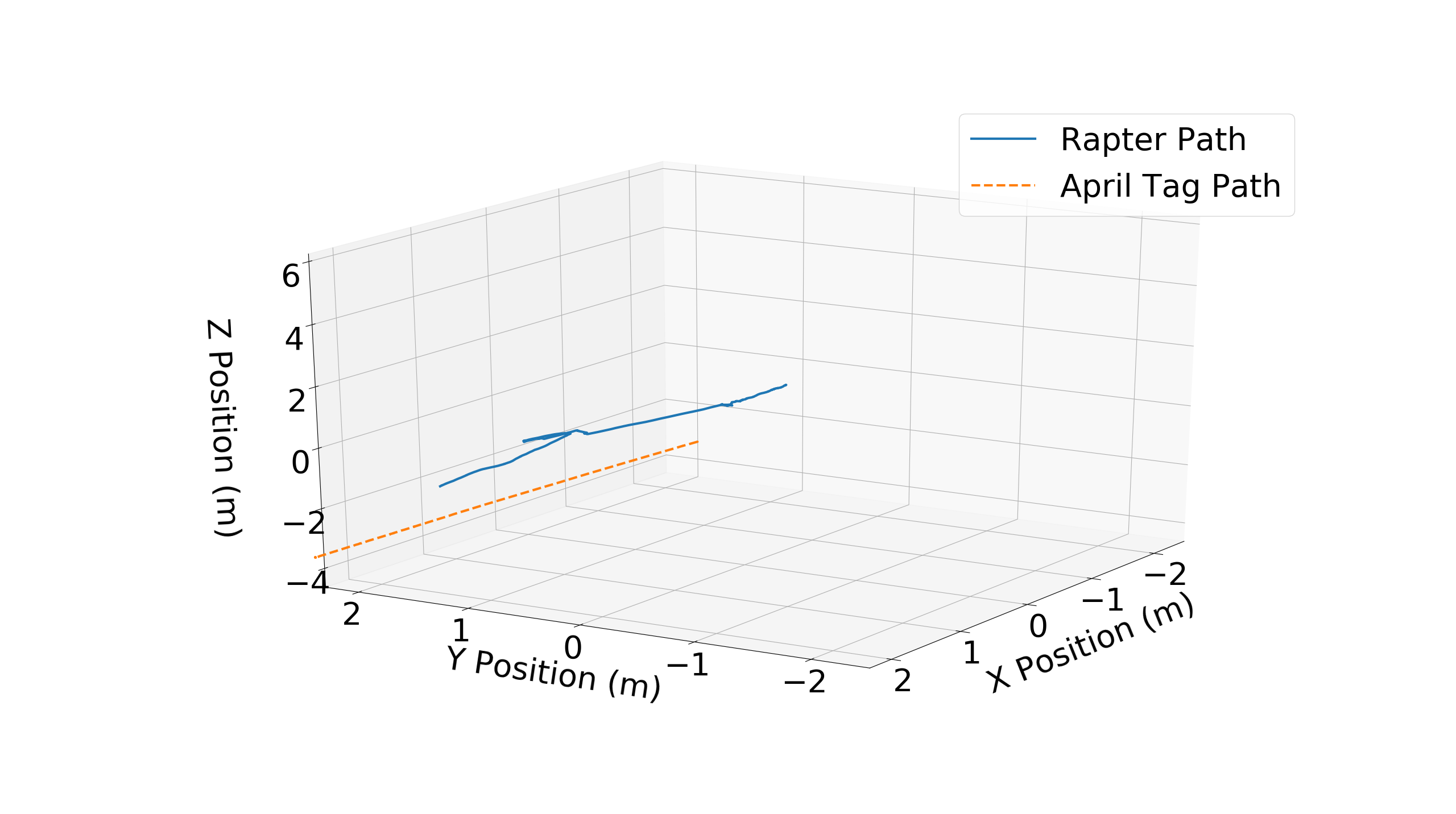}
    \caption{Rapter Trajectory with Turtle Trajectory using DINO w/obstruction}
    \label{fig:turtle_fan_dino_o}

    \includegraphics[width=0.75\linewidth]{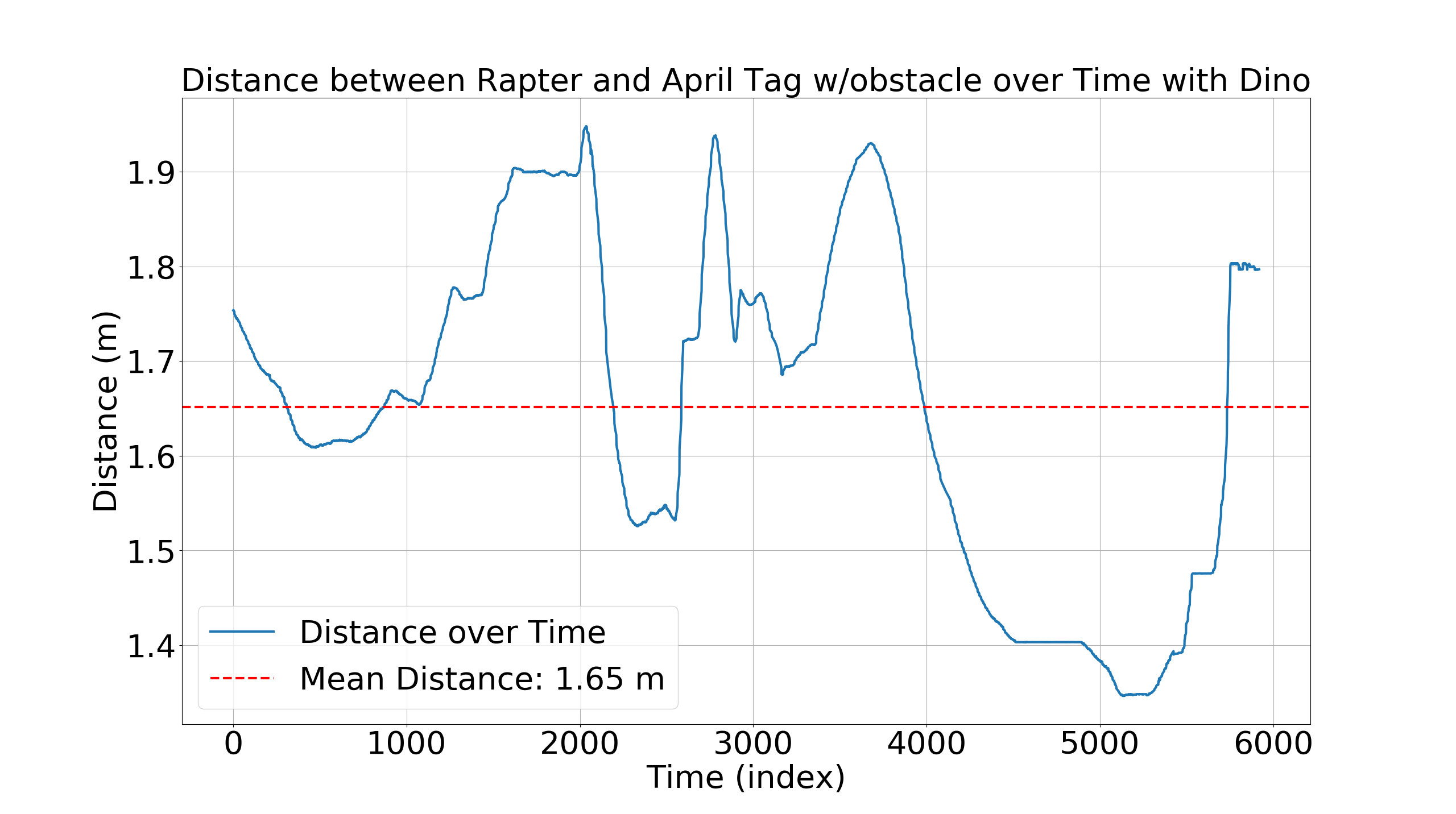}
    \caption{DTW(Dynamic Time Warping) of Rapter Trajectory with Turtle Trajectory using DINO w/obstruction}
    \label{fig:turtle_fan_dino_o_dtw}
    
\end{figure}

\subsection{Multimodal Input Queries}
The implemented one-shot tracking system was evaluated using multi-modal user input queries. The results from sample trials, where the input queries were either a bounding box around the target or an image template of the target, are presented in Figures \ref{fig:Dino_pipeline} and \ref{fig:bbox_pipeline}, respectively. It was observed that when input queries were based on real-time bounding boxes, the tracking performance was slightly compromised compared to scenarios where the input query was an image template of the target. This disparity can be attributed to potential inaccuracies in human-drawn bounding boxes, leading to erroneous segmentation and suboptimal feature extraction, thereby adversely affecting the overall tracking performance.

These offsets were significantly reduced when DINO was introduced to detect and track the features. In Figure \ref{fig:apriltag_fan_dino_no}, it is evident that the Rapter's trajectory is more stable compared to previous figures. This is corroborated by the calculated DTW for this run, which shows a mean DTW of 0.67 meters, indicating the drone had an offset of only 0.67 meters.

To further verify DINO's capabilities, we switched the target from an AprilTag back to a toy turtle and provided an image template of the turtle. DINO effortlessly detected the features and provided similar results to those obtained with the AprilTag. As shown in Figure \ref{fig:turtle_fan_dino_no_dtw}, the mean DTW is 0.62 meters, a remarkably low offset considering the specifications of the Rapter's tracking camera.

\subsection{Comparative Analysis: VOXL2 Apriltag Detection vs. FAN with Bounding Boxes vs FAN with DINO}
As part of our analysis, we aimed to compare the accuracy and reliability of our case studies and test the VOXL2 AprilTag detection method. This method utilizes position control based on VOXL's AprilTag pose estimation tracking. We then compared its performance with our FAN (Follow Anything) models, which include the Bounding Box (BB) and DINO modalities, using the same AprilTag.

As shown in Figure \ref{fig:Apriltag tracking w/voxl_to_mpa_ros2 method}{}, the VOXL-based AprilTag tracking, jittery at the beginning, later proceeds to track the AprilTag smoothly. In Figure \ref{fig:Apriltag tracking w/voxl_to_mpa_ros2 method dtw}, it can be seen that the mean DTW for VOXL-based AprilTag detection sums up to 0.80 meters, which is better than our FAN with bounding boxes. However, the DINO modality outperformed the others, achieving the lowest DTW of 0.67 meters.

\subsection{Tracking against obstruction}
To test the robustness of the Rapter's multi-modality, we conducted several test runs introducing occlusion to the target, tracking it while gradually occluding it from one end of the setup and re-emerging it from the other, ensuring the entire setup was within the Rapter's field of view (FOV). While testing on a wide dataset of frames, the Rapter could easily re-detect the target using DINO but failed to re-detect the targets on its own when they re-emerged. To test if it could detect any features, we assisted the Rapter by switching to "position mode" mid-run and flew it slightly closer to the target until it could re-detect it, which the Rapter successfully did. After numerous runs and analyses, we concluded that the re-detection algorithm itself is robust, but limited hardware and external conditions, which will be discussed later, inhibited the drone from re-detecting the target on its own. In Figures \ref{fig:apriltag_fan_dino_o} and \ref{fig:turtle_fan_dino_o}, a peculiar pattern in the Rapter's trajectory indicates where it lost visuals on the target.

\section{Conclusion}
In conclusion, this research project endeavors to address the limitations of existing robotic systems for object detection and tracking by designing and implementing a novel system that leverages state-of-the-art pre-trained models to accurately detect and track target objects through multimodal queries. By employing a custom-built UAV and integrating advanced algorithms, we have aimed to achieve precise and adaptable object tracking capabilities, surpassing the constraints of closed-set systems and enhancing usability through intuitive interaction methods.

Drawing inspiration from prior work, we have tailored and extended methodologies to suit our objectives, focusing on compatibility with our custom-built drone and developing a custom high-level algorithm for efficient object tracking. Through rigorous evaluation in simulated and real-world scenarios, we have demonstrated the robustness and efficacy of our detection and tracking models, particularly in handling obstructions and multi-modal input queries.

Our contributions include the development of a custom evaluation pipeline, a proportional-based high-level controller, and a ROS2-based implementation, enhancing integration and operational flexibility. Additionally, we have introduced the utilization of Discrete Time Warping (DTW) for trajectory evaluation, providing a systematic method for assessing the quality and effectiveness of the tracking algorithm.

\section{Future Work}
Future work involves several enhancements to improve the system's performance and expand its capabilities. The tracking algorithm can be refined by replacing the Proportional (P) controller with a Proportional-Integral-Derivative (PID) controller, which will provide more precise and stable tracking by addressing steady-state errors and improving response to dynamic changes. Additionally, integrating the "text" modality using the CLIP algorithm will broaden the range of input modalities, enhancing user interaction. Exploring alternative modalities, such as voice commands or gesture recognition, will further increase the system's versatility. Extensive real-world testing is necessary to evaluate robustness and adaptability under various conditions, including different lighting, weather, and complex backgrounds. Optimizing computational efficiency through model compression techniques and hardware acceleration will ensure effective real-time operation on resource-constrained platforms. Extending the system to support multi-target tracking will enhance its applicability in surveillance and crowd monitoring. Developing a more user-friendly interface with intuitive input methods, real-time feedback, and customizable settings will improve user interaction. Finally, integrating the system into collaborative multi-UAV setups for tasks like cooperative tracking, search and rescue, and large-scale environmental monitoring will expand its utility and effectiveness in complex scenarios.

\section{Acknowledgements}
We want to express our deepest gratitude to Professor Joseph Conroy for his invaluable guidance and support throughout the ENAE 788M course. His expertise and encouragement have been instrumental in the successful completion of this project. We sincerely appreciate his dedication to our academic growth and his continuous encouragement to explore innovative solutions. Thank you, Professor Conroy, for providing us with the knowledge and tools necessary to excel in our research endeavors. 


{\small
\bibliographystyle{unsrt}
\bibliography{ref.bib}
}

\end{document}